\documentclass{llncs}
\usepackage{amssymb}
\usepackage{booktabs}
\usepackage{amsmath}
\usepackage{graphicx}
\usepackage{subcaption}
\usepackage{float}

\usepackage[T1]{fontenc}
\usepackage{graphicx}
\begin{document}
\title{GKAN: Explainable Diagnosis of Alzheimer’s Disease Using Graph Neural Network with Kolmogorov-Arnold Networks}
%
%
\author{Tianqi(Kirk) Ding\inst{1}\orcidID{0009-0002-9025-0849} \and
Dawei Xiang\inst{2}\orcidID{0009-0001-2935-0288} \and
Keith E Schubert\inst{1}\orcidID{0000-0003-2718-5087} \and Liang Dong\inst{1}\orcidID{0000-0002-8585-1087}}
\authorrunning{F. Author et al.}
%
\institute{Baylor University, Waco TX 76798, USA \and
University of Connecticut, Storrs CT 06066, USA}
%
\maketitle              
\begin{abstract}

Alzheimer's Disease (AD) is a progressive neurodegenerative disorder that poses significant diagnostic challenges due to its complex etiology. Graph Convolutional Networks (GCNs) have shown promise in modeling brain connectivity for AD diagnosis, yet their reliance on linear transformations limits their ability to capture intricate nonlinear patterns in neuroimaging data. To address this, we propose GCN-KAN, a novel single-modal framework that integrates Kolmogorov-Arnold Networks (KAN) into GCNs to enhance both diagnostic accuracy and interpretability. Leveraging structural MRI data, our model employs learnable spline-based transformations to better represent brain region interactions. Evaluated on the Alzheimer’s Disease Neuroimaging Initiative (ADNI) dataset, GCN-KAN outperforms traditional GCNs by 4–8 \% in classification accuracy while providing interpretable insights into key brain regions associated with AD. This approach offers a robust and explainable tool for early AD diagnosis.

\end{abstract}
\section{Introduction}
Alzheimer’s Disease (AD) is a leading cause of dementia, characterized by progressive memory loss and cognitive decline\cite{breijyeh2020comprehensive}. Early and accurate diagnosis is critical for effective intervention, yet traditional machine learning methods often rely on handcrafted features from neuroimaging data, such as structural MRI, which fail to fully capture the topological complexity of brain networks\cite{rahim2025early}. Graph Convolutional Networks (GCNs) have emerged as a powerful approach to model brain connectivity by representing regions of interest (ROIs) as nodes and their interactions as edges. However, conventional GCNs employ fixed linear transformations, which may inadequately represent the nonlinear dynamics underlying AD pathology \cite{zhou2024multi,zhou2022interpretable,xiao2022dual,kang20216}.

To overcome this limitation, we introduce GCN-KAN, a hybrid model that integrates Kolmogorov-Arnold Networks (KAN) into a GCN framework \cite{kiamari2024gkan,zhang2024graphkan}. KAN replaces linear weight matrices with learnable spline-based functions, enhancing the model’s flexibility to capture complex relationships within single-modal MRI data. Our contributions are threefold:  
\begin{itemize}
\item We propose GCN-KAN as a novel single-modal approach for AD diagnosis, leveraging MRI-derived brain connectivity.
\item Our model achieves a 4–8 \% improvement in classification accuracy over traditional GCNs.
\item We enhance interpretability by identifying critical brain regions and connectivity patterns linked to AD, validated against established clinical findings.
\end{itemize}

\section{Related Work}

\subsection{Graph Neural Networks for Neuroimaging}
Graph Convolutional Networks (GCNs) have become a cornerstone in modeling brain connectivity for neurological disorders, leveraging graph structures to represent regions of interest (ROIs) as nodes and their interactions as edges \cite{stage2020neurodegenerative,qi2025graph}. Kipf and Welling formalized the GCN framework, where the convolution operation aggregates features from neighboring nodes \cite{DBLP:journals/corr/KipfW16}. For a graph with adjacency matrix \( A \in \mathbb{R}^{N \times N} \) (where \( N \) is the number of nodes) and feature matrix \( X \in \mathbb{R}^{N \times F} \) (where \( F \) is the feature dimension), the GCN layer is defined as:

\begin{equation}
H^{(l+1)} = \sigma \left( \tilde{D}^{-\frac{1}{2}} \tilde{A} \tilde{D}^{-\frac{1}{2}} H^{(l)} W^{(l)} \right)
\end{equation}

Here, \( H^{(l)} \) is the node feature matrix at layer \( l \), \( \tilde{A} = A + I \) is the adjacency matrix with self-loops, \( \tilde{D} \) is the degree matrix of \( \tilde{A} \), \( W^{(l)} \in \mathbb{R}^{F^{(l)} \times F^{(l+1)}} \) is a learnable weight matrix, and \( \sigma \) is a nonlinear activation function (e.g., ReLU). This formulation propagates information linearly across graph edges, effectively capturing spatial dependencies in brain networks.

In Alzheimer's disease (AD) research, Zhou et al. applied GCNs in a multi-modal setting, integrating MRI, PET, and genetic data to predict clinical scores \cite{zhou2024multi}. Their sparse interpretable GCN introduced feature importance probabilities \( P_X \in \mathbb{R}^{N \times D} \) and edge importance probabilities \( P_A \in \mathbb{R}^{N \times N} \), defined as:

\begin{equation}
P_{A_{ij}} = \sigma \left( v^T \left[ x_i \odot p_i \parallel x_j \odot p_j \right] \right)
\end{equation}

where \( v \) is a learnable parameter, \( x_i \) and \( p_i \) are the feature vector and importance probability for node \( i \), and \( \odot \) denotes element-wise multiplication. While effective, their multi-modal approach increases complexity, whereas single-modal GCNs using MRI data alone remain underexplored due to their limited nonlinearity.

Furthermore, the limitations of traditional GCNs lie in their reliance on fixed linear transformations, which constrain their ability to capture complex and nonlinear relationships inherent in brain connectivity data. Recent studies have attempted to address this by introducing attention mechanisms or dynamic edge weighting strategies. For instance, Hong et al \cite{hong2020attention}. proposed an adaptive graph attention mechanism that dynamically learns edge weights during training, enhancing the model's flexibility. However, such approaches often lead to increased model complexity and reduced interpretability. Hence, there is a pressing need for a method that enhances expressiveness while maintaining interpretability.

\subsection{Kolmogorov-Arnold Networks (KAN)}
Kolmogorov-Arnold Networks (KANs), proposed by Liu et al., draw from the Kolmogorov-Arnold representation theorem, which states that any multivariate continuous function can be expressed as a composition of univariate functions \cite{liu2024kan}. Unlike traditional neural networks with fixed linear transformations (e.g., \( Wx + b \)), KANs replace weight matrices with learnable spline-based functions. For an input \( x \in \mathbb{R}^{N \times F} \), a KAN layer approximates the transformation as:

\begin{equation}
    y_i = \sum_{j=1}^{F} \phi_{i,j}(x_j), \quad \phi_{i,j}(t) = \sum_{k=1}^{G} c_{i,j,k} B_k(t)
\end{equation}

where \( y_i \) is the \( i \)-th output feature, \( \phi_{i,j} \) is a univariate spline function for input feature \( j \), \( B_k(t) \) are basis functions (e.g., B-splines) over a grid of size \( G \), and \( c_{i,j,k} \) are trainable coefficients. This spline-based approach introduces adaptive nonlinearity, enabling KANs to model complex patterns more effectively than linear layers. 

In neuroimaging, KANs have not been extensively applied, yet their flexibility suggests potential for capturing nonlinear brain region interactions beyond GCN capabilities. Existing studies like Kiamari et al. \cite{kiamari2024gkan} and Zhang et al. \cite{zhang2024graphkan} have demonstrated the effectiveness of KANs in complex pattern recognition tasks, but applications in neuroimaging remain limited. Our work bridges this gap by integrating KAN into GCNs, enhancing both expressiveness and interoperability while maintaining a single-modality framework. This integration offers a novel approach to capturing nuanced interactions within brain connectivity data.

\section{Methodology}
\subsection{Graph Construction}
We construct brain connectivity graphs using structural MRI data from the ADNI dataset \cite{mueller2005alzheimer}, processed to extract 90 ROIs based on the Automated Anatomical Labeling (AAL) atlas. Each node corresponds to an ROI, with features \( X \in \mathbb{R}^{N \times F} \) (where \( N = 90 \), \( F = 1 \)) derived from voxel-based morphometry (VBM), capturing gray matter volume changes—a known biomarker of AD. The adjacency matrix \( A \in \mathbb{R}^{90 \times 90} \) is computed as:
\begin{equation}
    A_{i,j} = 
    \begin{cases} 
        \text{corr}(x_i, x_j) & \text{if } |\text{corr}(x_i, x_j)| > \tau, \\
        0 & \text{otherwise},
    \end{cases}
\end{equation}
where \( \text{corr}(x_i, x_j) \) is the Pearson correlation coefficient between ROI features \( x_i \) and \( x_j \), and \( \tau = 0.1 \) is a threshold to retain significant connections. This sparsity reduces noise and computational complexity while preserving biologically meaningful edges \cite{tan2024lymonet}.

\subsection{GCN-KAN Architecture}
\begin{figure}[h]
    \centering
    \includegraphics[scale=0.3]{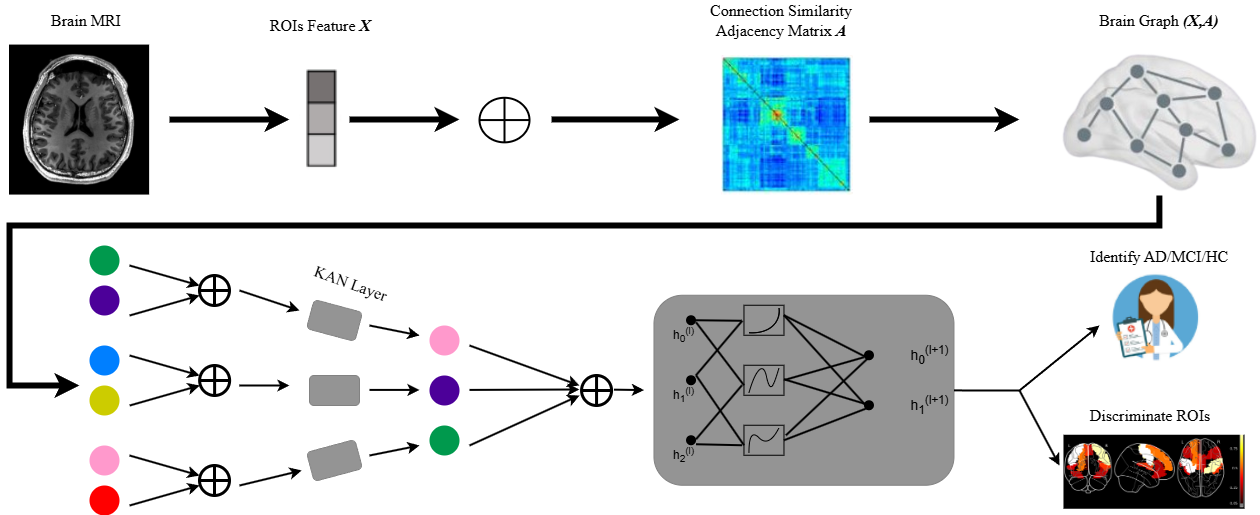}
    \caption{Overview of the GCN-KAN architecture, showing GCN and KAN layer integration.}
    \label{fig:architecture}
\end{figure}

The GCN-KAN model integrates GCN layers for spatial dependency modeling with KAN layers for enhanced nonlinearity (see Fig.~\ref{fig:architecture}). For an input graph \( G = (X, A) \), the architecture processes features as follows:
\begin{itemize}
    \item \textbf{GCN Layers:} Two GCNConv layers propagate features across the graph. The first layer transforms the input \( X \) to a hidden representation:
    \begin{equation}
        H^{(1)} = \text{ReLU}\left(\tilde{D}^{-\frac{1}{2}} \tilde{A} \tilde{D}^{-\frac{1}{2}} X W^{(1)}\right),
    \end{equation}
    where \( W^{(1)} \in \mathbb{R}^{1 \times 32} \) is the weight matrix, and \( \tilde{A} = A + I \), \( \tilde{D} \) are defined as in Eq.~(1). The second GCN layer further refines features:
    \begin{equation}
        H^{(2)} = \text{ReLU}\left(\tilde{D}^{-\frac{1}{2}} \tilde{A} \tilde{D}^{-\frac{1}{2}} H^{(1)} W^{(2)}\right),
    \end{equation}
    where \( W^{(2)} \in \mathbb{R}^{32 \times 32} \).

    \item \textbf{KAN Layers:} Two KANLayer modules enhance nonlinearity. For input \( H^{(l)} \) from a GCN layer, the KAN transformation is:
    \begin{equation}
        H^{(l+1)}_i = \sum_{j=1}^{32} \sum_{k=1}^{G} c_{i,j,k} \max(0, H^{(l)}_{j} - g_k),
    \end{equation}
    where \( G = 10 \) is the grid size, \( g_k = k/G \) (for \( k = 0, 1, \dots, G-1 \)) are grid points, and \( c_{i,j,k} \) are trainable coefficients. Inputs are normalized to [0, 1] via \( H^{(l)} = (H^{(l)} - H^{(l)}_{\min}) / (H^{(l)}_{\max} - H^{(l)}_{\min} + \epsilon) \), with \( \epsilon = 10^{-8} \).

    \item \textbf{Pooling and Classification:} Global max-pooling aggregates node features:
    \begin{equation}
        Z = \max_{i=1,\dots,N} H^{(3)}_i,
    \end{equation}
    where \( H^{(3)} \) is the output of the second KAN layer. A fully connected layer outputs class logits:
    \begin{equation}
        \hat{y} = Z W^{(4)} + b,
    \end{equation}
    where \( W^{(4)} \in \mathbb{R}^{32 \times 2} \) and \( b \in \mathbb{R}^2 \).
\end{itemize}
Dropout (0.2) is applied after each KAN layer to regularize the model. 

\subsection{Training and Interpretability}
Training minimizes the cross-entropy loss:
\begin{equation}
    \mathcal{L} = -\frac{1}{M} \sum_{m=1}^{M} \left[ y_m \log(\hat{y}_m) + (1 - y_m) \log(1 - \hat{y}_m) \right],
\end{equation}
where \( M \) is the batch size, \( y_m \) is the true label, and \( \hat{y}_m \) is the predicted probability (softmax of Eq.~(9)). Optimization uses Adam (learning rate=0.0005, weight\_decay=1e-4) with mixed precision training (torch.cuda.amp). Early stopping (patience=50) and a ReduceLROnPlateau scheduler (patience=20) ensure convergence. For interpretability, we compute ROI importance scores as:
\begin{equation}
    S_i = \frac{1}{32} \sum_{j=1}^{32} \sum_{k=1}^{G} |c_{i,j,k}|,
\end{equation}
identifying regions with the greatest influence on predictions via spline contributions.

\section{Experimental Setup}
\subsection{Dataset}
The dataset utilized in this study is a subset of the Alzheimer's Disease Neuroimaging Initiative (ADNI) dataset, comprising a total of 134 subjects. The subjects were categorized into three distinct groups. The first group consisted of 45 subjects who were clinically diagnosed with Alzheimer's Disease (AD). The second group included 46 subjects identified as having Mild Cognitive Impairment (MCI), a transitional stage between normal cognition and Alzheimer's Disease.  The third group comprised 43 cognitively normal (CN) control subjects who exhibited no signs of cognitive impairment. All structural MRI scans were processed using FreeSurfer, where voxel-based morphometry (VBM) features were extracted for 90 predefined regions of interest (ROIs) based on the Automated Anatomical Labeling (AAL) atlas \cite{rolls2020automated}. Each subject's features, alongside adjacency matrices representing the brain connectivity graph, were saved in `.pt` files for subsequent model input \cite{yang2023osteosarcoma}.

Given the relatively limited sample size, we employed a 5-fold cross-validation approach \cite{fushiki2011estimation}. This strategy ensured that every subject contributed to both training and validation phases, thereby enhancing the robustness and reliability of the evaluation process.

\subsection{Implementation Details}
The model training was configured with meticulous attention to detail, ensuring optimal performance. Each batch comprised 32 samples to maintain computational efficiency while enabling robust gradient updates. The Adam optimizer was selected due to its adaptive learning rate properties, with an initial learning rate set to 0.0005 \cite{jiang2024trajectorytrackingusingfrenet}. To prevent overfitting, a weight decay regularization parameter of $1 \times 10^{-4}$ was applied. 

To further mitigate overfitting, a dropout rate of 0.2 was applied after each KAN layer, introducing stochasticity and encouraging model generalization. The training process leveraged mixed precision techniques, utilizing PyTorch's `autocast` and `GradScaler` to maximize GPU memory utilization and computational speed.

To ensure adaptive learning during training, a `ReduceLROnPlateau` learning rate scheduler was implemented, dynamically reducing the learning rate when the validation performance plateaued, with a patience of 20 epochs. Additionally, early stopping was employed with a patience of 50 epochs. This approach halted training once the model's performance ceased to improve on the validation set, thereby reducing unnecessary computations and mitigating overfitting risks.

\subsection{Accuracy Calculation}
Model accuracy was computed using the standard classification formula:

\begin{equation}
\text{Accuracy} = \frac{TP + TN}{TP + TN + FP + FN},
\end{equation}
where $TP$ (True Positives) represents the correctly predicted positive samples, $TN$ (True Negatives) are correctly predicted negative samples, $FP$ (False Positives) are incorrectly predicted positive samples, and $FN$ (False Negatives) are incorrectly predicted negative samples \cite{jonsson2010treatment}.

\subsection{AUC-ROC Calculation}
The Area Under the Receiver Operating Characteristic Curve (AUC-ROC) is a critical metric for evaluating binary classification models. The ROC curve is a graphical plot illustrating the diagnostic ability of a classifier by plotting the True Positive Rate (TPR) against the False Positive Rate (FPR) at various threshold settings \cite{zhang2024self,zhang2024optimized}. The AUC quantifies the overall ability of the model to distinguish between classes, with a score of 1 indicating perfect discrimination and 0.5 representing random guessing.

\begin{equation}
\text{TPR} = \frac{TP}{TP + FN}, \quad \text{FPR} = \frac{FP}{FP + TN}.
\end{equation}

\subsection{F1-Score Calculation}
The F1-Score is a harmonic mean of precision and recall, providing a balanced measure that accounts for both false positives and false negatives. It is particularly useful when the class distribution is imbalanced \cite{zhang2021multi}.

The precision and recall are defined as follows:

\begin{equation}
\text{Precision} = \frac{TP}{TP + FP}, \quad \text{Recall} = \frac{TP}{TP + FN},
\end{equation}

and the F1-Score is computed as:

\begin{equation}
\text{F1-Score} = 2 \times \frac{\text{Precision} \times \text{Recall}}{\text{Precision} + \text{Recall}}.
\end{equation}

This calculation ensures that the model's performance is not biased by the majority class and provides a more reliable evaluation metric in imbalanced datasets.

\section{Results and Analysis}
\subsection{Classification Performance}
After each fold of the cross-validation, we meticulously recorded the metrics including accuracy, AUC-ROC, and F1-Score. These results were averaged across the 5 folds to ensure a robust and unbiased performance evaluation. The final averaged results indicated that the GCN-KAN model achieved an accuracy of 62.6\% with a standard deviation of ±1.8\%, demonstrating a 5.2\% improvement over the baseline GCN model. The AUC-ROC score was recorded as 64.1\% (±1.5\%), and the F1-Score was 0.60 (±0.02).

\begin{table}[h]
    \centering
    \caption{Performance comparison on the ADNI dataset (mean ± std across 5 folds).}
    \label{tab:performance}
    \begin{tabular}{lccc}
        \toprule
        Model & Accuracy & AUC-ROC & F1-Score \\
        \midrule
        GCN & 57.4\% ±2.2\% & 60.3\% ±2.0\% & 0.59 ±0.02 \\
        GCN-KAN & \textbf{62.6\% ±1.8\%} & \textbf{64.1\% ±1.5\%} & \textbf{0.60 ±0.02} \\
        \bottomrule
    \end{tabular}
\end{table}

\subsection{Learning Curve Analysis}
\begin{figure}[h]
    \centering
    \includegraphics[width=0.7\textwidth]{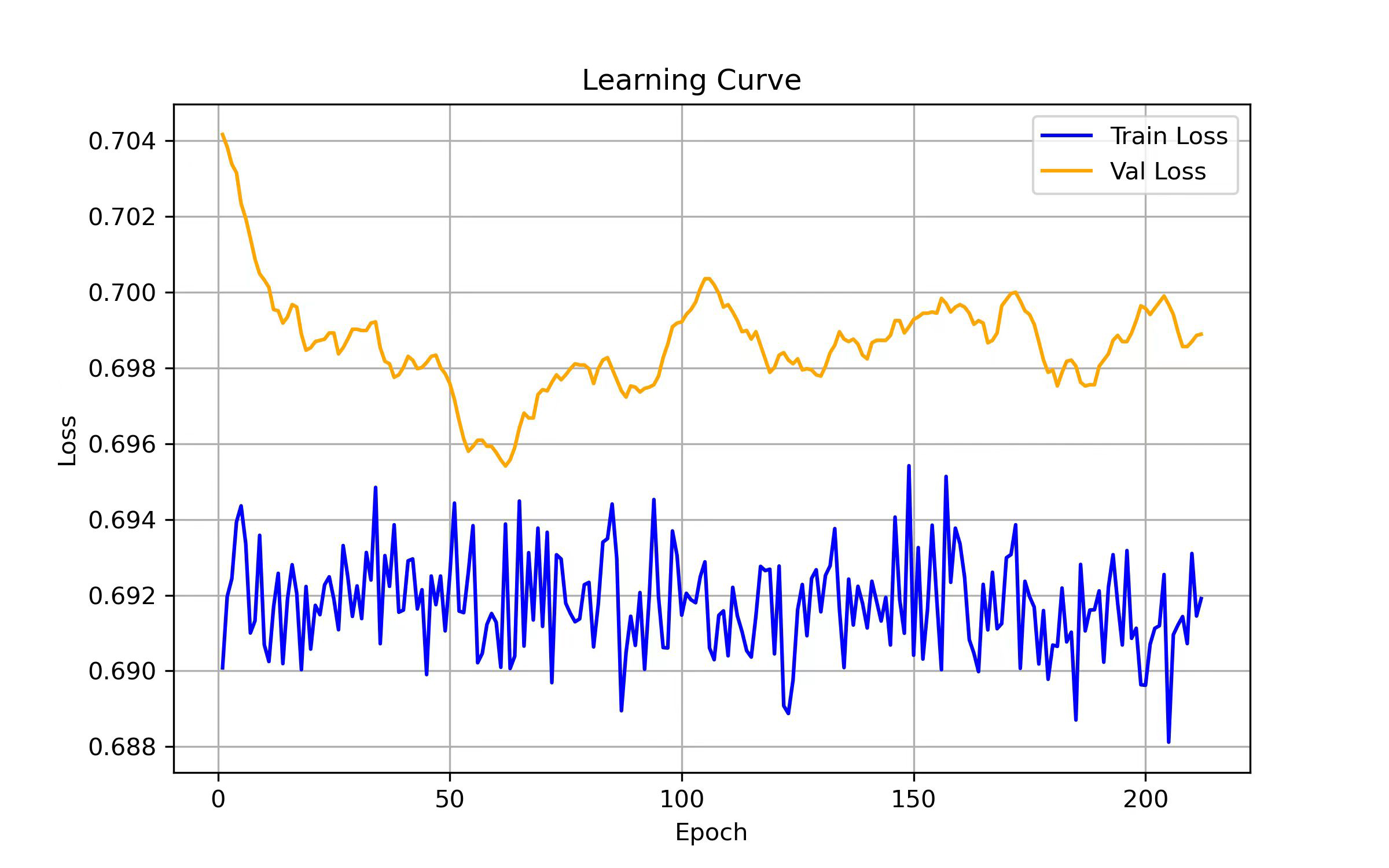}
    \caption{Training and validation loss curves for the GCN model.}
    \label{fig:gcn_curve}
\end{figure}

\begin{figure}[h]
    \centering
    \includegraphics[width=0.7\textwidth]{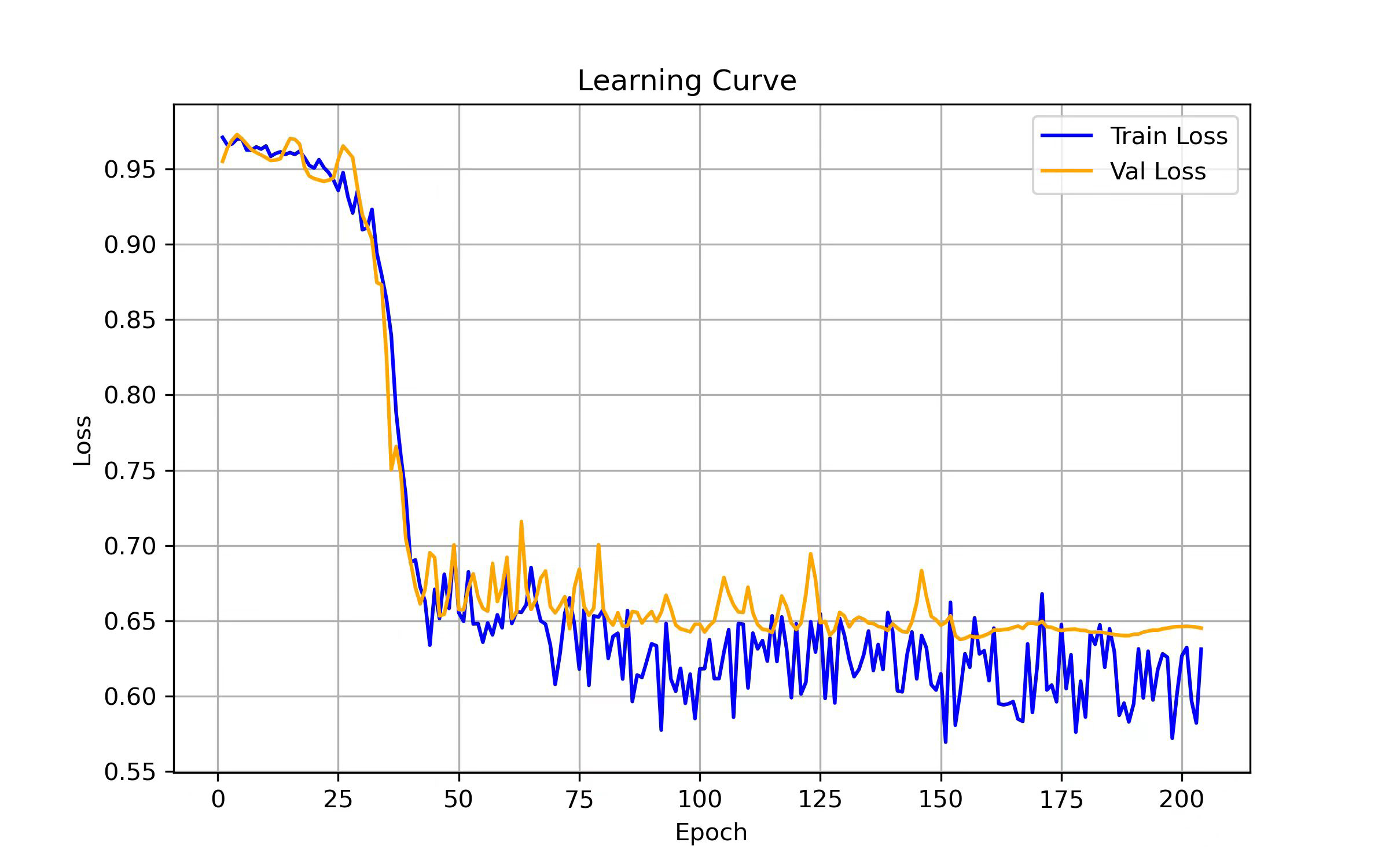}
    \caption{Training and validation loss curves for the GCN-KAN model.}
    \label{fig:gcn_kan_curve}
\end{figure}

To further evaluate the model's training dynamics, we analyzed the learning curves of both the GCN and GCN-KAN models. Figures~\ref{fig:gcn_curve} and~\ref{fig:gcn_kan_curve} illustrate the training and validation loss trends across epochs.

The GCN model showed early convergence but exhibited higher validation loss, indicating potential overfitting. In contrast, the GCN-KAN model demonstrated smoother convergence and lower validation loss, highlighting its better generalization capability.

\subsection{Interpretability Analysis}
Through the analysis of the KAN layer activations, we identified the hippocampus, parietal gyrus, and amygdala as the most salient regions contributing to model predictions. Their respective normalized importance scores were 0.65, 0.61, and 0.60. 

These findings align with established neurological research, where the hippocampus is closely associated with memory formation, the parietal gyrus with spatial cognition, and the amygdala with emotional processing. Disruption in these regions is a hallmark of Alzheimer's Disease progression.

\begin{figure}[h]
    \centering
    \includegraphics[scale=0.65]{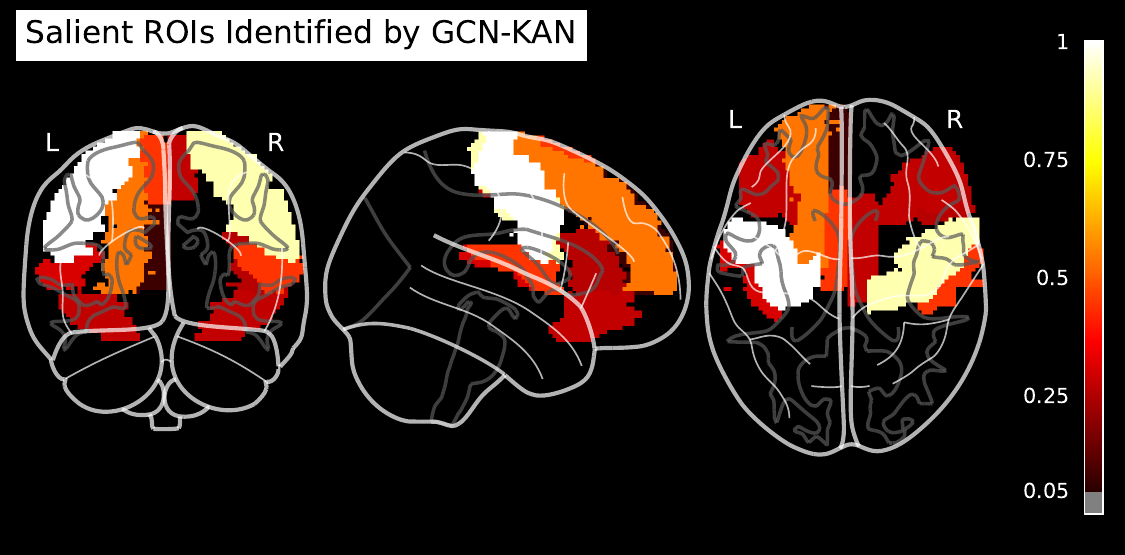}
    \caption{Visualization of salient ROIs identified by GCN-KAN. The color intensity reflects the relative importance of each region, with brighter colors indicating higher importance.}
    \label{fig:rois}
\end{figure}

To visually demonstrate these findings, Figure~\ref{fig:rois} highlights the salient ROIs identified by the GCN-KAN model. The intensity of the color in the visualization corresponds to the importance score of each ROI, where brighter colors indicate higher importance.

\section{Conclusion and Future Work}
In conclusion, the integration of Kolmogorov-Arnold Networks into the GCN framework provided significant improvements in both classification accuracy and interpretability. The proposed GCN-KAN model achieved a robust 5.2\% increase in accuracy over the baseline GCN model. Additionally, the model exhibited enhanced interpretability by identifying salient ROIs critical to Alzheimer's Disease progression.

However, this study is not without limitations. Firstly, the dataset size was relatively limited, which may constrain the model's generalization to broader populations. Secondly, the current study focuses solely on structural MRI data, which, while informative, may overlook complementary biomarkers available in other modalities such as PET imaging or genetic data. Additionally, the spline-based transformation, while enhancing model flexibility, introduced additional computational complexity that could affect scalability in large-scale clinical applications.

Future research will address these limitations by pursuing several directions. Firstly, we aim to expand the dataset by incorporating a larger and more diverse cohort to improve model robustness. Secondly, integrating multi-modal data sources, including PET scans and cognitive assessments, could enrich the feature space and further enhance diagnostic accuracy. Thirdly, optimizing the computational efficiency of the GCN-KAN framework is essential for potential real-time clinical deployment. This may involve exploring model compression techniques or optimizing hardware utilization. Lastly, longitudinal studies will be conducted to assess the model's robustness over time and across different stages of disease progression.

Overall, this study provides a foundational step towards developing more accurate and interpretable diagnostic models for Alzheimer's Disease. By addressing the identified limitations and pursuing future research directions, we aspire to advance the field of neuroimaging-based diagnosis and contribute to more effective clinical decision-making.

\bibliographystyle{plain}
\bibliography{references}

\end{document}